\documentclass[10pt,twocolumn,letterpaper]{article}

\usepackage{cvpr}              

\def\paperID{16940} 
\def\confName{CVPR}
\def\confYear{2025}

\title{Noise is an Efficient Learner for Zero-Shot Vision-Language Models}

\author{Raza Imam\quad Asif Hanif\quad Jian Zhang\quad Khaled Waleed Dawoud\quad \\ Yova Kementchedjhieva \quad Mohammad Yaqub\\
Mohamed bin Zayed University of Artificial Intelligence (MBZUAI), UAE\\
{\tt\small \{firstname.lastname\}@mbzuai.ac.ae}
\\ \small {\url{https://github.com/Razaimam45/TNT}}
}

\usepackage{graphicx}
\usepackage{amsmath}
\usepackage{amssymb}
\usepackage{booktabs}

\usepackage{caption}   %
\usepackage{listings}
\usepackage{xcolor}

\usepackage{algorithm}
\usepackage{algpseudocode}
\usepackage{listings}

\newcommand{\x}{\bm{\mathrm{x}}}

\newcommand{\txt}{\bm{\mathrm{t}}}

\definecolor{codegray}{rgb}{0.5,0.5,0.5}

\lstdefinestyle{Pytorch}{
    language         = Python,
    backgroundcolor  = \color{white},
    basicstyle = \fontsize{8.0pt}{10pt}\selectfont\ttfamily\bfseries,
    columns          = fullflexible,
    breaklines       = true,
    captionpos       = b,
    commentstyle     = \fontsize{4pt}{4pt}\color{codeblue},
    keywordstyle     = \fontsize{4pt}{4pt}\color{codekw},
    morekeywords     = {noise, confidence\_filter, vc\_loss, entropy},
    frame=None,
}

\definecolor{cvprblue}{rgb}{0.21,0.49,0.74}
\usepackage[pagebackref,breaklinks,colorlinks]{hyperref}
\usepackage{orcidlink}
\usepackage{colortbl}
\usepackage{float}
\usepackage{bbding}
\usepackage{pifont}
\usepackage{multirow}
\usepackage{amsmath}
\usepackage{bm}
\usepackage{amsfonts}
\usepackage{enumitem}
\usepackage{caption}
\usepackage{wrapfig}

\usepackage[capitalize]{cleveref}
\crefname{section}{Sec.}{Secs.}
\Crefname{section}{Section}{Sections}
\Crefname{table}{Table}{Tables}
\crefname{table}{Tab.}{Tabs.}

\definecolor{citecolor}{RGB}{0, 113, 188}
\definecolor{myforestgreen}{RGB}{34, 200, 34}
\definecolor{firebrick}{rgb}{0.7, 0.13, 0.13}
\definecolor{darkpastelgreen}{rgb}{0.01, 0.75, 0.24}
\definecolor{deepskyblue}{rgb}{0.0, 0.75, 1.0}
\definecolor{mypink2}{rgb}{.99,.96,.98}
\definecolor{mypink1}{rgb}{.99,.93,.98}
\definecolor{mypink}{rgb}{.99,.90,.98}
\definecolor{mygray}{rgb}{.95,.95,.95}
\definecolor{lv14}{rgb}{0.5,0.5,0.5}
\definecolor{tabvline}{HTML}{a8a495}
\definecolor{prompt_blue}{HTML}{1f78b4}
\definecolor{prompt_red}{HTML}{d45c43}
\definecolor{green_im}{rgb}{0.0, 0.5, 0.0}

\definecolor{customblue}{HTML}{6aaed6} 
\definecolor{customgreen}{HTML}{89bf91} 
\definecolor{customrocket}{HTML}{f58860} 

\definecolor{codeblue}{rgb}{0.25, 0.5, 0.5}
\definecolor{codekw}{rgb}{0.35, 0.35, 0.75}

\newcommand{\stdvuno}[1]{\footnotesize{\color{black}(#1)} {\color{black}}}
\newcommand{\stdvun}[1]{\footnotesize\textcolor{black}{(#1)} \textcolor{myforestgreen}{$\blacktriangle$}}
\newcommand{\stdvuru}[1]{\footnotesize\textcolor{black}{({#1})} \textcolor{red}{$\blacktriangledown$}}

\begin{document}

\maketitle

\begin{abstract}

    Recently, test-time adaptation has garnered attention as a method for tuning models without labeled data. The conventional modus operandi for adapting pre-trained vision-language models (VLMs) during test-time primarily focuses on tuning learnable prompts; however, this approach overlooks potential distribution shifts in the visual representations themselves. In this work, we address this limitation by introducing \textbf{Test-Time Noise Tuning (TNT)}, a novel method for handling unpredictable shifts in the visual space. TNT leverages, for the first time, a \textbf{noise adaptation} strategy that optimizes learnable noise directly in the visual input space, enabling adaptive feature learning from a single test sample. We further introduce a novel approach for \textit{inter-view representation alignment} by explicitly enforcing coherence in embedding distances, ensuring consistent feature representations across views. Combined with scaled logits and confident view selection at inference, TNT substantially enhances VLM generalization and calibration, achieving average gains of +7.38\% on natural distributions benchmark and +0.80\% on cross-dataset evaluations over zero-shot CLIP. These improvements lay a strong foundation for adaptive out-of-distribution handling. 

  \end{abstract}


\section{Introduction}
Vision-language models (VLMs) have been shown to successfully perform various downstream tasks in a zero-shot fashion, which eliminates the need for creating task-specific training data and storing multiple models. The ability of VLMs to generalize on open-world problems, however, degrades as real-world data shifts away from the distribution that the models were trained on. Test-Time Adaptation (TTA) has thus emerged as a critical approach to enhance model robustness while maintaining the advantages of zero-shot learning  \cite{sun2020test, wang2020tent}. TTA operates at inference time, leveraging unlabeled test data to dynamically adjust existing or newly added model parameters to the desired distribution.

Various forms of parametrization for TTA have been explored, both on the side of the image encoder and the text encoder of VLMs \cite{xiao2024beyond}. These include learning a soft prompt for the text encoder \cite{shu2022test, yooncctpt, murugesan2024robust}, adapting the batch norm layers of the VLM \cite{li2016revisitingbatchnormalizationpractical, zhang2022memotesttimerobustness}, learning LoRA layers in model components \cite{imam2024testtimelowrankadaptation}, or even updating whole components of the model, such as the vision encoder \cite{zhao2023test}.
Amidst an abundance of TTA studies, one form of parametrization that remains unexplored is \textit{noise}.

\begin{figure}[t]
    \centering
    \includegraphics[width=\linewidth]{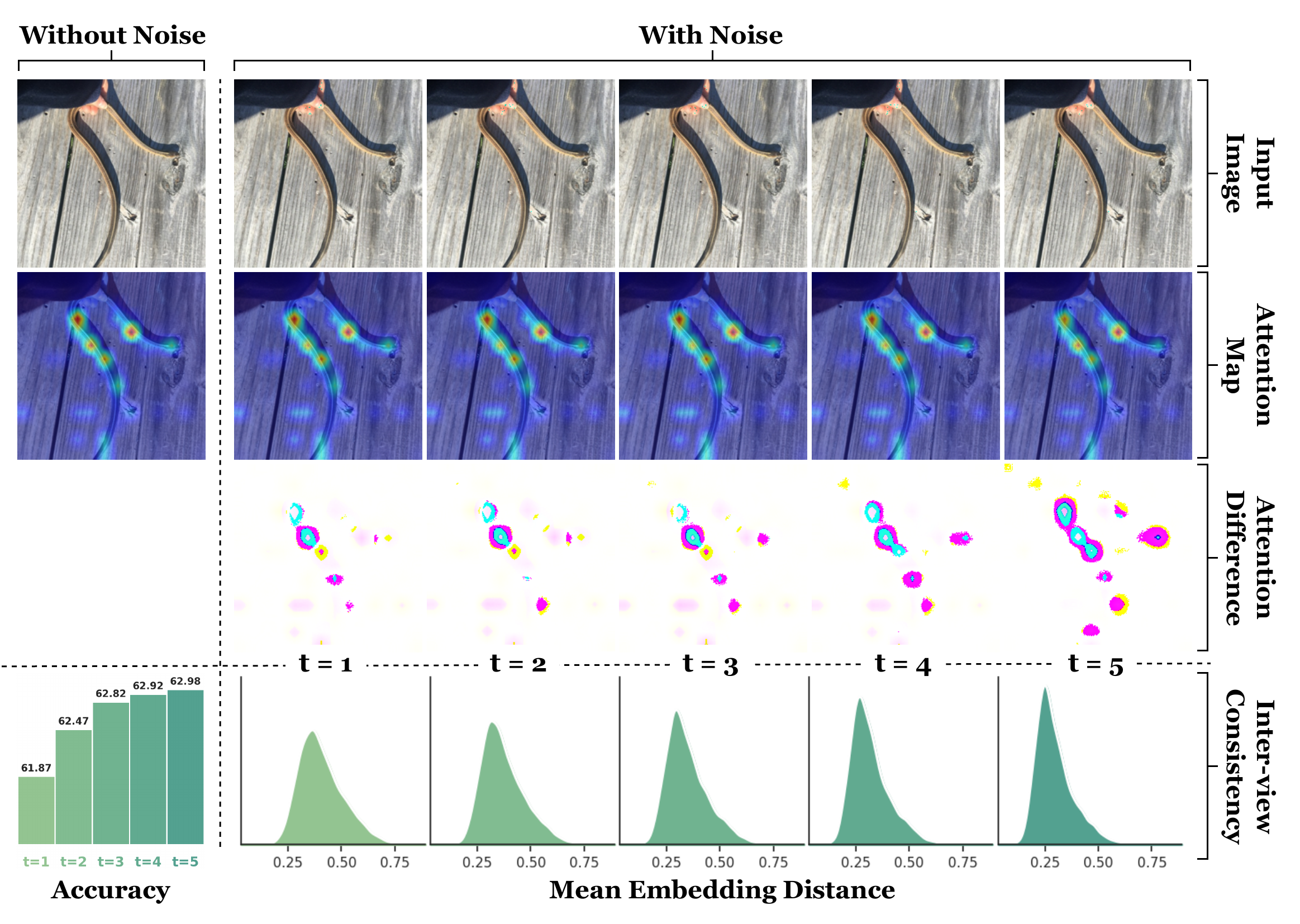}
    \vspace{-0.6cm}
    \caption{ As top-\( K \) augmented view embeddings grow more consistent with each optimization step \( t \), the attention mechanism focuses on relevant regions, leading to improved accuracy.  \textit{Attention Difference} illustrates the absolute difference between the clean attention map and the noise-tuned attention map. CLIP zero-shot incorrectly classifies the original image as {\color{red}{amaga}}, while TNT correctly classifies the optimized image as {\color{myforestgreen}{garter snake}}.
    }
    \label{fig:concept}
    \vspace{-0.4cm}
\end{figure}

While noise is often seen as disruptive to machine learning models, it has also proved valuable in many ways. In generative models such as generative adversarial networks \cite{goodfellow2014generative}, variational autoencoders \cite{kingma2013auto}, and, more recently, diffusion models \cite{ho2020denoising, rezende2014stochastic, radford2015unsupervised}, \textit{random} noise has been used to initialize and guide output generation with impressive results. 
Building on findings that noise can enhance representation learning and improve robustness under varying conditions \cite{bengio2013representation, song2019generative}, recent works \cite{hanif2024baple, bai2024badclip} have applied learnable noise to examine adversarial vulnerabilities in VLMs using prompt learning setup. Inspired by this approach, we propose a TTA framework based on noise adaptation.

Our approach, dubbed Test-time Noise Tuning (TNT), relies on \textit{learnable noise}, applied over the augmented views of an input image to enhance regions relevant to its correct classification. The noise is sampled from a standard Gaussian distribution and optimized with a two-fold objective: (1) minimizing marginal entropy \cite{zhang2022memo}, an approach that proves effective when applied to input adaptation, as demonstrated previously in label adaptation \cite{shu2022test}; and (2) maximizing inter-view consistency, a novel objective introduced to promote consistency across representations of different augmented views of the input image. By ensuring that different augmentations of the image map to similar points in the embedding space, the model learns to focus on core, invariant features over superficial details (see an example in Figure~\ref{fig:concept}.) The learned noise is applied to both the input image and its augmentations in an enhanced inference procedure, with performance further boosted through temperature scaling \cite{murugesan2024robust,tu2024empirical}. 

In a comparative evaluation against seven strong TTA baselines on two established out-of-distribution benchmarks, our approach proves both more accurate and better calibrated, which is crucial to real-world applications.
Our method achieves this performance without considerable latency compared to other methods, and proves highly effective even with a limited parametrization budget. 
In summary, our main contributions are as follows:
\begin{itemize}[leftmargin=0.40cm, itemsep=0.10em]
    \item We propose \textbf{TNT}, a novel \textit{noise} adaptation strategy that optimizes VLM's vision encoder's input space by incorporating learnable noise at test time for a single test sample, enhancing model robustness to distributional shifts and improving out-of-distribution generalization. 
    \item We introduce an inter-view consistency loss for the noise tuning strategy that minimizes the distance between confident augmented views, fostering more aligned representations and reducing prediction uncertainty. This approach harmonizes the benefits of consistency loss and entropy minimization on top of sole noise adaptation.
    \item We demonstrate that TNT significantly improves VLM generalization, achieving state-of-the-art performance on a range of natural shift and cross-dataset benchmarks, all with reduced computational overhead.
\end{itemize}
\noindent To the best of our knowledge, we are the first to explore noise optimization for representation learning within VLMs, offering a novel TTA approach. Our findings on TNT highlight the potential of noise-adaptive architectures, encouraging further research into enhancing model robustness and generalization.

\section{Related Work}
\noindent\textbf{Zero-Shot VLM Generalization:}
Pre-trained on large image-text datasets in a self-supervised way, VLMs such as CLIP \cite{radford2021learning} and ALIGN \cite{jia2021scaling} have shown strong generalization capabilities. For example, CLIP's remarkable zero-shot transfer performance can be attributed to the diversity and scale of the data on which it was trained. Nonetheless, adapting them effectively to specific downstream tasks when data is scarce remains a challenge. One straightforward yet effective approach to enhance CLIP’s zero-shot performance on image classification is the use of \textit{soft prompts} \cite{zhou2022coop} which are \textit{learned} in a few-shot training setup. 



\vspace{0.2cm}
\noindent\textbf{Test-Time Optimization:}
Approaches such as TPT \cite{shu2022test} adjust prompts at test time to reduce entropy across augmented views of a single test sample, improving accuracy without additional training data. However, TPT does not address \textit{model calibration}, which is essential for uncertainty estimation.
To remedy this, C-TPT \cite{yooncctpt} improves calibration by optimizing prompt selection based on the dispersion of text features, eliminating the need for labeled data. Reinforcement learning with CLIP feedback (RLCF) \cite{zhaotestrlcf} further enhances generalization by offering continuous feedback during TTA, correcting predictions and preventing overconfidence associated with entropy minimization in TPT. Sample-wise Temperature Scaling (SaLS) \cite{murugesan2024robust} modifies temperature scaling during TTA on top of TPT to boost calibration, refining the model’s confidence. CALIP \cite{guo2023calip}, enhances CLIP’s zero-shot performance by incorporating an attention module that enables interaction between visual and textual features, all without requiring additional training or parameters.

\vspace{0.2cm}
\noindent\textbf{Noise-based Learning:} 
While learnable noise has not been explored for TTA, two recent works -- BadCLIP \cite{bai2024badclip} and BAPLE \cite{hanif2024baple} -- have used it to inject backdoor triggers into the image encoder's input within a few-shot training setup. These approaches introduce learnable noise during the prompt learning stage to compromise the VLM, demonstrating that simply adding noise can alter the model's behavior. This insight raises an intriguing question: \textit{could such noise be harnessed positively}?



\begin{figure*}[t]
    \centering
    \includegraphics[width=1.0\textwidth]{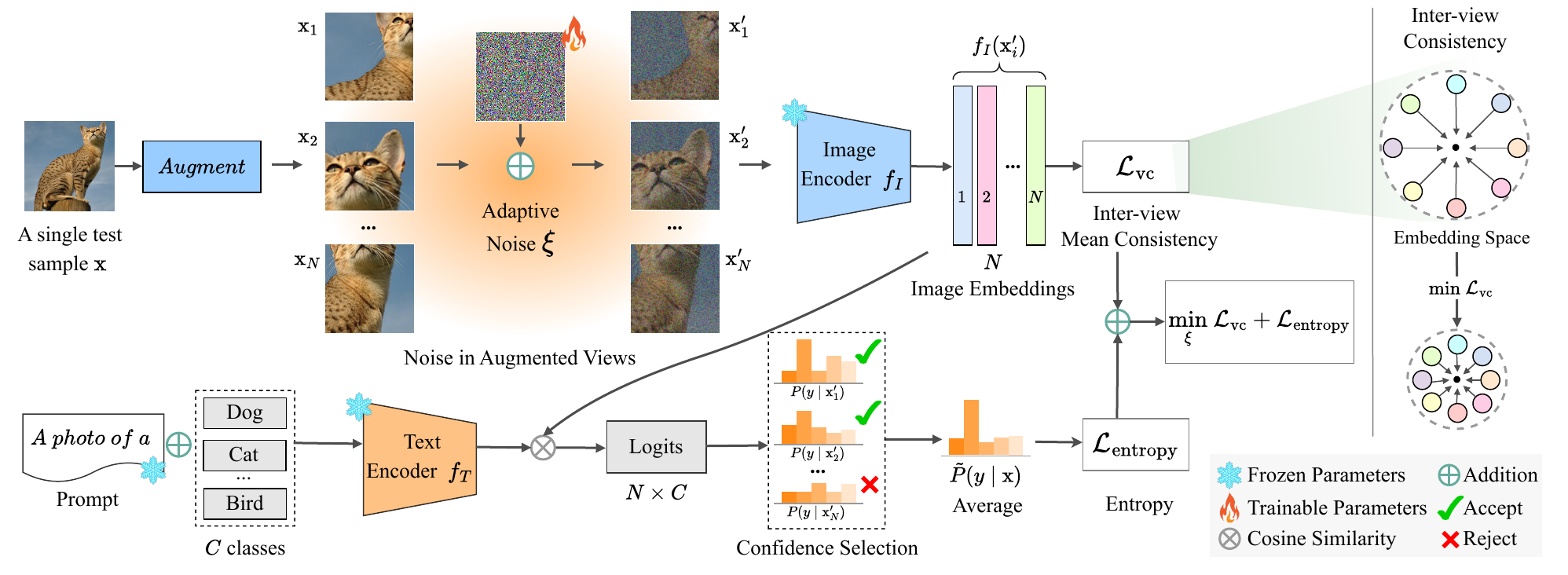}
    \vspace{-0.50cm}
    \caption{\textbf{Test-Time Noise Tuning (TNT)} (1) generates augmented views of a test image, (2) applies adaptive \textit{learnable} noise, and (3) computes logits and feature vectors for each view. (4) Top-\(K\) views are selected by confidence, with (5) entropy loss [Eq. \ref{eq:entropy}] enforcing confident predictions and (6) inter-view consistency loss [Eq. \ref{eq:Lvc_loss}] aligning feature representations. (7) The combined loss is backpropagated to iteratively refine the noise, enabling adaptive test-time noise tuning.}
    \label{fig:method}
    \vspace{-0.20cm}
\end{figure*}

\section{TNT: Test-Time Noise Tuning}

\subsection{Zero-shot Image Classification with VLMs}
Foundation VLMs, such as CLIP and ALIGN, have been shown to perform well on the task of image classification using a simple but effective zero-shot approach. Given an input image, \(\x\), and a set of class descriptions, \(\txt = \{t_1, t_2, \dots, t_c\}\) for a total of \(c\) classes, the prediction scores can be obtained as: 
\begin{equation}
\label{eq:logits}
f(\x,\txt) = \bigg\{\operatorname{sim}\big(f_{I}(\x), f_{T}(t_i)\big)\bigg\}_{i=1}^{c}
\end{equation}
where \( f_I \) denotes the image encoder of the VLM, \( f_T \), the text encoder and \(\operatorname{sim(\dot)}\) is cosine-similarity. For brevity, we hereafter drop \(\txt\) and denote the scores as \(f(\x) \in  \mathbb{R}^{c} \). In this work, we rely on this general framework for zero-shot image classification and experiment specifically with CLIP, without loss of generalization. Although current VLMs exhibit impressive generalization across visual domains and object classes, we aim to improve their performance further through \textit{noise optimization} in the context of TTA.

\subsection{Noise Optimization}
Unlike traditional fine-tuning, which risks domain-specific biases, prompt tuning refines the input text context to preserve the VLM's pre-trained features, enhancing its ability to retrieve relevant knowledge with precision. However, relying solely on text optimization may overlook essential visual details. We propose enhancing the image encoder's input with \textit{learnable} noise, allowing the model to capture subtle features, align with text prompts, and improve performance on varied or noisy images by preserving critical visual cues. Our approach keeps the model's weights unchanged, preserving its zero-shot capabilities. The workflow of our approach is outlined below.

\vspace{0.2cm}
\noindent\textbf{Noise in Augmented Views:} 
Consider an input image \( \x \in \mathbb{R}^{C \times H \times W} \) at test time, where \( C \), \( H \), and \( W \) denote channels, height, and width, respectively. To generate diverse views, \( N \) \textit{augmented} versions of \( \x \), represented by \( (\x_1, \x_2, \dots, \x_N) \), are created using random transformations. The key feature of our approach is a \textit{learnable} noise map \( \xi \in \mathbb{R}^{C \times H \times W}\), which is added to each augmented view and tuned through test-time adaptation.  Noise values, \( \xi \),  are constrained to \([-\epsilon, +\epsilon]\), where \(\epsilon\) is the perturbation budget. The perturbed \(i_\text{th}\) augmented view is obtained as follows:
\begin{equation}
\x^{\prime}_i = \operatorname{clamp}(\x_i + \xi,~ 0,~1)
\end{equation}
where \( \operatorname{clamp}(\cdot) \) constrains the input values within the valid range \([0, 1]\). At each optimization step, the noise values are iteratively updated to reduce model uncertainty following the objectives described below.  

\vspace{0.2cm}
\noindent\textbf{Entropy Loss:} The model, \( f(\cdot) \), generates logits (unnormalized prediction scores) for each augmented view, denoted as \( f(\mathbf{x}'_i) \) for \( i \in \{1, 2, \dots, N\} \). Following the approach in \cite{shu2022test}, we retain only the high-confidence views, selecting the top-\( K \) views with lowest self-entropy.\footnote{Notice that the dynamic percentile threshold implemented in \cite{shu2022test} effectively implements a top-K selection process as well.} Using these top-\( K \) views, we compute the marginal entropy loss, denoted by \(\mathcal{L}_{\text{entropy}}(\cdot)\), as follows:
\begin{equation}
\mathcal{L}_{\text{entropy}} =  \mathcal{H}\bigg(\frac{1}{K}\sum_{k \in \mathcal{K}}\operatorname{softmax}(f(\mathbf{x}'_k))\bigg)
\label{eq:entropy}
\end{equation}
where \(\mathcal{H}(\cdot)\) computes the entropy of the average probability distribution. This distribution is obtained by first applying \(\operatorname{softmax}(\cdot)\) to each logit \( f(\mathbf{x}'_k) \) and then averaging the resulting probabilities across the top-\( K \) high-confidence views. Here, \(\mathcal{K}\) represents the set of indices corresponding to these top-\( K \) views.

\vspace{0.2cm}
\noindent\textbf{Inter-view Consistency Loss:}  
To ensure consistency among multiple augmented views, we introduce \textit{inter-view consistency} loss minimization objective to penalize large embedding variations among high-confidence views. Let \( f_{I}(\x^{\prime}_i) \in \mathbb{R}^{d} \) represent the embedding of  perturbed \(i_\text{th}\) augmented view \( \x^{\prime}_i \). We calculate pairwise Euclidean \( (\ell_2) \) distance among embeddings of the top-\( K \) selected views and define the \textit{inter-view consistency} loss, denoted by \( \mathcal{L}_{\text{vc}} \), as follows:
\begin{equation}
\mathcal{L}_{\text{vc}} = \sum_{i \in \mathcal{K}} \sum_{j \in \mathcal{K}} \big\| f_{I}(\x^{\prime}_i) - f_{I}(\x^{\prime}_j) \big\|_2
\label{eq:Lvc_loss}
\end{equation}
where \(i\neq j\). \( \mathcal{L}_{\text{vc}} \) loss, combined with noise perturbation, penalizes high variance in representations, encouraging the model to maintain stable and consistent feature embeddings for the selected confident views. 

\vspace{0.2cm}
\noindent\textbf{Noise Tuning:}  The final loss \(\mathcal{L}\) combines the entropy and inter-view consistency losses i.e. \(\mathcal{L} = \alpha \mathcal{L}_{\text{entropy}} + \beta \mathcal{L}_{\text{vc}}\), where \(\alpha\) and \(\beta\) control the weights of the respective losses. The learnable noise is adapted by minimizing \(\mathcal{L}\) as follows:
\begin{equation}
\label{eq:tnt_objective}
\underset{ \xi }{\operatorname{minimize}}~~~~ \alpha \mathcal{L}_{\text{entropy}} + \beta \mathcal{L}_{\text{vc}}
\end{equation}
This objective encourages both high-confidence predictions (through entropy minimization) and consistent embeddings across views (via inter-view distance minimization), thereby enhancing model robustness and calibration under distribution shifts. To update the noise \( \xi \) in each iteration, we compute the gradient of the loss \( \mathcal{L} \) and update \( \xi \) as follows:
\begin{equation}
\xi \leftarrow \xi - \gamma \cdot \operatorname{sign}(\nabla_\xi \mathcal{L})
\end{equation}
where \( \gamma \) is the learning rate, and \( \operatorname{sign}(\cdot) \) denotes the element-wise sign function \cite{goodfellow2015explainingharnessingadversarialexamples}. After each update step, the noise \( \xi \) values are clamped to the interval \( [-\epsilon, +\epsilon] \). Algorithm \ref{alg:tnt} provides a high-level overview of the noise update process in TNT.

\vspace{0.2cm}
\noindent\textbf{Inference:} 
After the noise adaptation phase, during inference, we add the \textit{learned} noise \( \xi \) to each of the \(N\) augmented views of the image \(\mathbf{x}\). Based on self-entropy, we then select the top-\(K\) most confident views, denoted as \( \{\mathbf{x}_k + \xi\}_{k \in \mathcal{K}} \). Using Equation \ref{eq:logits}, we obtain the logits  for each of the perturbed top-\(K\) views and compute the final probability distribution \(\mathbf{p}\) as follows:
\begin{equation}
\mathbf{p} = \frac{1}{K} \sum_{k \in \mathcal{K}} \operatorname{softmax}_{\tau}\big(f(\mathbf{x}_k + \xi)\big)
\label{eq:inference}
\end{equation}
where \(\tau\) is the temperature parameter in the \(\operatorname{softmax}\) function which scales the obtained logits. The predicted label is then obtained by taking the \(\operatorname{argmax}\) of the probability distribution i.e. 
\begin{equation}
\hat{y} = \underset{ i\in \{1,2,\dots,c\} }{\operatorname{argmax}}~~p_i
\end{equation}
where \(p_i\) is the probability of \(i_{\text{th}}\) class in distribution \(\mathbf{p}\). 

\begin{algorithm}[t]
\caption{PyTorch style Pseudocode for TNT}
\label{alg:tnt}
\vspace{-1.ex}
\begin{lstlisting}[style=Pytorch,escapeinside={(@}{@)}]
# image = single test image 
# model = pre-trained VLM
# eps = perturbation budget for noise
# lr = learning rate to update noise
# top_k = number of top-K views
# t = number of steps
def TNT(model, image, top_k, t):
  noise = torch.randn(image.shape)
  noise = noise.clamp(-eps,eps)
  # Get N augmented views of test image
  images = augment(image) # (N,C,H,W)
  for step in range(t)
     # Add noise to augmented views of test image
     # and get logits and views' feature vectors
     logits, images_feats = model(images + noise)
     # Get indices of top-K views 
     k_indices = confidence_filter(logits, top_k)
     # Compute entropy loss on top-K views
     loss_e = entropy(logits[k_indices])
     # Compute inter-view consistency loss
     loss_vc = vc_loss(images_feats[k_indices]) 
     loss = loss_e + loss_vc
     loss.backward()
     # Update the learnable noise
     noise = noise - torch.sign(noise.grad)*lr
     noise = noise.clamp(-eps,eps)
  # Inference after noise adaptation
  logits, _ = model(images + noise)
  # Apply temperature scaling on top-K views
  probs = (logits[k_indices]/tau).softmax()
  predicted_label =  argmax(probs.mean(dim=0))
  return predicted_label
\end{lstlisting}
\vspace{-1.ex}
\end{algorithm}

{\renewcommand{\arraystretch}{1.0}
\begin{table*}[t]
\caption{\textbf{Top-1 accuracy} $\%$ of state-of-the-art baselines, where \textbf{OOD Avg.} indicates the OOD average results and {\cellcolor{mygray}{$bs.$}} indicates the baseline, \ie, CLIP-ViT-B-16. The arrow ${\color{myforestgreen}\blacktriangle}$ and ${\color{red}\blacktriangledown}$ indicate \textbf{improvements} and \textbf{decrements} compared to the CLIP method, \ie, CLIP-ViT-B/16. RLCF* denotes RLCF with all visual encoder parameters trainable. TNT* denotes proposed method with hand crafted prompts while TNT denotes CoOp initialized prompts. \textbf{Bold} indicates best performance, \underline{Underline} indicates second-best.}
\centering
\resizebox{\textwidth}{!}{%
\begin{tabular}{l|c|cccc|c|c}
\hline
\textbf{Method} $\downarrow$ & \textbf{ImageNet} & \textbf{ImageNet-A} & \textbf{ImageNet-V} & \textbf{ImageNet-R} & \textbf{ImageNet-K} & \textbf{Average} & \textbf{OOD Avg.} \\ 
\cmidrule(r){1-1} \cmidrule(lr){2-2} \cmidrule(lr){3-3} \cmidrule(lr){4-4} \cmidrule(lr){5-5} \cmidrule(lr){6-6} \cmidrule(l){7-7} \cmidrule(l){8-8}

CLIP-ViT-B/16 & 67.41\stdvuno{$bs.$} & 47.85\stdvuno{$bs.$} & 60.89\stdvuno{$bs.$} & 73.99\stdvuno{$bs.$} & 46.10\stdvuno{$bs.$} & 59.25\stdvuno{$bs.$} & 57.21\stdvuno{$bs.$} \\

CoOp & 68.63\stdvun{1.22} & 50.25\stdvun{2.40} & \underline{64.95}\stdvun{4.06} & 75.70\stdvun{1.71} & \underline{48.26}\stdvun{2.16} & 61.56\stdvun{2.31} & 59.79\stdvun{2.58} \\

TPT${\color{cyan}_{\text{NIPS '22}}}$ & 69.86\stdvun{2.45} & 54.25\stdvun{6.40} & 63.19\stdvun{2.30} & 76.90\stdvun{2.91} & 47.45\stdvun{1.35} & 62.33\stdvun{3.08} & 60.45\stdvun{3.24} \\

CALIP${\color{cyan}_{\text{AAAI '23}}}$ & 66.74\stdvuru{0.67} & 47.76\stdvuru{0.09} & 60.76\stdvuru{0.13} & 73.99\stdvuru{0.00} & 46.12\stdvun{0.02} & 59.07\stdvuru{0.18} & 57.16\stdvuru{0.05} \\

C-TPT${\color{cyan}_{\text{ICLR '24}}}$ & 68.56\stdvun{1.15} & 50.67\stdvun{2.82} & 61.56\stdvun{0.67} & 75.32\stdvun{1.33} & 46.84\stdvun{0.74} & 60.59\stdvun{1.34} & 58.60\stdvun{1.39} \\

SaLS${\color{cyan}_{\text{ECCV '24}}}$ & 69.67\stdvun{2.26} & 54.53\stdvun{6.68} & 63.22\stdvun{2.33} & 76.88\stdvun{2.89} & 47.51\stdvun{1.41} & 62.36\stdvun{3.11} & 60.54\stdvun{3.33} \\

RLCF${\color{cyan}_{\text{ICLR '24}}}$ & 69.36\stdvun{1.95} & 54.08\stdvun{6.23} & 62.71\stdvun{1.82} & 76.82\stdvun{2.83} & 47.33\stdvun{1.23} & 62.06\stdvun{2.81} & 60.24\stdvun{3.03} \\

RLCF*${\color{cyan}_{\text{ICLR '24}}}$ & 70.14\stdvun{2.73} & 59.24\stdvun{11.39} & 64.55\stdvun{3.66} & \underline{77.13}\stdvun{3.14} & 48.50\stdvun{2.40} & 63.91\stdvun{4.66} & 62.36\stdvun{5.15} \\

\rowcolor[HTML]{E2EFDA} 
TNT* & \underline{70.27}\stdvun{2.86} & \underline{61.87}\stdvun{14.02} & 63.64\stdvun{2.75} & 76.96\stdvun{2.97} & 48.03\stdvun{1.93} & \underline{64.15}\stdvun{4.90} & \underline{62.63}\stdvun{5.42} \\

\rowcolor[HTML]{E2EFDA} 
\(\text{TNT}\) & \textbf{72.06}\stdvun{4.65} & \textbf{63.93}\stdvun{16.08} & \textbf{66.64}\stdvun{5.75} & \textbf{78.61}\stdvun{4.62} & \textbf{49.16}\stdvun{3.06} & \textbf{66.08}\stdvun{6.83} & \textbf{64.59}\stdvun{7.38} \\ \hline



\end{tabular}
}
\label{tab:natural_shift}
\vspace{-0.3cm}
\end{table*}}

\section{Experiments and Results}
\subsection{Experimental Setup}

\noindent\textbf{Datasets:}
We conduct experiments on a diverse range of benchmark datasets to assess the performance and robustness of our method, specifically testing its out-of-domain generalization across different domains. The selected datasets include ImageNet-A \cite{hendrycks2021natural}, ImageNet-V2 \cite{recht2019imagenet}, ImageNet-R \cite{hendrycks2021many} and ImageNet-Sketch (denoted as ImageNet-K) \cite{wang2019learning}, which have been considered as out-of-distribution (OOD) data for ImageNet to assess model robustness under different conditions and distributions.

For cross-domain generalization, following \cite{shu2022test}, we include Flowers102 \cite{nilsback2008automated}, DTD \cite{cimpoi2014describing}, Pets \cite{parkhi2012cats}, UCF \cite{soomro2012ucf101}, and Caltech101 \cite{fei2004learning}. These datasets were chosen to analyze the model's ability to distinguish subtle differences between similar classes. Additionally, we include Aircraft \cite{maji2013fine}, EuroSAT \cite{helber2019eurosat}, Cars \cite{krause20133d}, Food \cite{bossard2014food}, and SUN397 \cite{xiao2010sun} to further test the model’s adaptability across distinct categories, including aerial images, satellite images, and object-centric as well as scene-centric datasets.

\vspace{0.1cm}
\noindent\textbf{Implementation Details:}
We initialize the noise \(\xi \in \mathbb{R}^{3\times 224\times224}\) by sampling from a standard Gaussian distribution and set \(\epsilon=1/255\). This noise is applied to \(N=64\) images, consisting of the original image and 63 augmented views, which are generated through random resized crops and horizontal flips of the original image. Noise is updated with a learning rate of \(1\mathrm{e-}3\) across all datasets. For temperature scaling, we use a constant value of $\tau = 7\mathrm{e-}3$ across all settings.
We use fixed prompts in two configurations: first, a hand-crafted prompt ("a photo of a \{CLASS\}"), referred to as TNT*; and second, 4-shot context weights obtained using CoOp \cite{zhou2022coop} on ImageNet, referred to as TNT.  All experiments are conducted on a single NVIDIA A6000 48GB GPU.

\vspace{0.1cm}
\noindent\textbf{Baselines:}
We evaluate a total of seven zero-shot baselines to assess the effectiveness of our approach. In addition to CLIP zero-shot \cite{radford2021learning} and CLIP with the CoOp pretrained soft prompt \cite{zhou2022learning}, we \textit{reproduce} several recently published test-time adaptation methods: TPT \cite{shu2022test},  CALIP \cite{guo2023calip}, C-TPT \cite{yooncctpt}, SaLS \cite{murugesan2024robust} and RLCF \cite{zhao2023test}.
All methods are reproduced on our system with a single update step and the same consistent backbones to ensure fair comparisons. Specifically, we use the same backbone for both the teacher and student models in RLCF.

\subsection{TNT Results}
\noindent\textbf{Natural Distribution Shifts:} 
Table \ref{tab:natural_shift} compares our method with seven baselines using a ViT-B/16 backbone, including zero-shot CLIP. Our simpler variant, TNT*, which uses a hand-crafted prompt, achieves the highest average performance across four OOD datasets and in-domain on ImageNet. This demonstrates that adapting input features through noise learning with marginal entropy and inter-view distance minimization effectively enhances classification accuracy. We achieve this \textit{without having to modify the large vision encoder} itself, as was done in \(\text{RLCF}^{*}\), yet we score substantially higher at a lower computational cost. While TNT* shows a minor in-domain improvement over \(\text{TPT}\) (\(< 0.5\) points), it achieves a substantial OOD gain (around 2 points), with ImageNet-A benefiting notably from noise adaptation. Further analysis of trainable parameters and text vs. image adaptation is detailed in \S\ref{comp_analysis}.

Our stronger variant, \(\text{TNT}\), uses a CoOp-initialized prompt in the text encoder, yielding an additional ~2 points in average performance both in- and out-of-domain. Interestingly, CoOp alone ranks poorly among the baselines, underperforming TNT by nearly 5 points. We observe an interesting synergy where the CoOp in itself has limited capacity but proves very effective as an initialization technique in TNT, enabling noise to adapt visual features and improving classification performance. In \S\ref{ablations}, we show that further tuning this prompt calibrates model predictions but does not improve accuracy.

\vspace{0.1cm}
\noindent\textbf{Cross-Dataset Generalization:} 
Table \ref{tab:cross_data} shows TNT's cross-dataset evaluation across ten datasets, following \cite{shu2022test}. TNT outperforms all baselines with an average accuracy of 64.48\% and achieving top generalization on seven out of ten datasets. TNT* also performs well, averaging 64.07\% and surpassing CLIP, CoOp, and TPT on several datasets. Notably, TNT* excels over TPT on OxfordPets, UCF, Aircraft, and StanfordCars. TNT's use of learnable noise enhances the visual feature space, capturing subtle distinctions across datasets and ensuring consistent, class-specific embeddings for improved zero-shot generalization, all without modifying the pre-trained model. The minimization of inter-view mean distance ensures consistent embeddings, enhancing the model's ability to differentiate class-specific details that \(\text{TPT}\) \cite{shu2022test} and \(\text{RLCF}\) \cite{zhaotestrlcf} miss.

{\renewcommand{\arraystretch}{1.0}
\begin{table*}[t]

\caption{\textbf{Top-1 accuracy} $\%$ of state-of-the-art baselines, where \textbf{Average} indicates average accuracies of the \textit{Cross-Datasets Generalization}. The arrow ${\color{myforestgreen}\blacktriangle}$ and ${\color{red}\blacktriangledown}$ indicate \textbf{improvements} and \textbf{decrements} compared to the CLIP-ViT-B/16. TNT* denotes the proposed method with hand-crafted prompts while  \(\text{TNT}\) denotes CoOp initialized prompts. \textbf{Bold} indicates best performance, \underline{Underline} indicates second-best.}

\centering
\resizebox{\textwidth}{!}{%
\begin{tabular}{l|cccccc} \hline
\textbf{Method} $\downarrow$ & ~\textbf{Flower102}\cite{nilsback2008automated}~ & ~\textbf{DTD}\cite{cimpoi2014describing}~ & ~\textbf{OxfordPets}\cite{parkhi2012cats}~ & ~\textbf{UCF}\cite{soomro2012ucf101}~ & ~\textbf{Caltech101}\cite{fei2004learning}~ & ~\textbf{Aircraft}\cite{maji2013fine}~ \\ 
\cmidrule(r){1-1} \cmidrule(lr){2-2} \cmidrule(r){3-3} \cmidrule(r){4-4} \cmidrule(r){5-5} \cmidrule(l){6-6} \cmidrule(l){7-7}

CLIP-ViT-B/16 & 67.92\stdvuno{$bs.$} & 44.27\stdvuno{$bs.$} & 88.28\stdvuno{$bs.$} & 65.21\stdvuno{$bs.$} & 93.39\stdvuno{$bs.$} & 23.82\stdvuno{$bs.$} \\

CoOp & \underline{68.82}\stdvun{0.90} & 39.24\stdvuru{5.03} & \underline{89.53}\stdvun{1.25} & \underline{68.5}1\stdvun{3.30} & 90.71\stdvuru{2.68} & 20.01\stdvuru{3.81} \\

TPT${\color{cyan}_{\text{NIPS '22}}}$ & 68.58\stdvun{0.66} & \textbf{45.27}\stdvun{1.00} & 86.29\stdvuru{2.00} & 67.62\stdvun{2.41} & 93.35\stdvuru{0.04} & 22.02\stdvuru{1.80} \\

CALIP${\color{cyan}_{\text{AAAI '23}}}$ & 67.64\stdvuru{0.28} & 44.44\stdvun{0.17} & 87.82\stdvuru{0.46} & 64.05\stdvuru{1.16} & 93.27\stdvuru{0.12} & \underline{24.12}\stdvun{0.30} \\

C-TPT${\color{cyan}_{\text{ICLR '24}}}$ & \textbf{70.24}\stdvun{2.32} & 44.74\stdvun{0.47} & 87.73\stdvuru{0.55} & 64.37\stdvuru{0.84} & 92.41\stdvuru{0.98} & 23.76\stdvuru{0.06} \\

SaLS${\color{cyan}_{\text{ECCV '24}}}$ & 68.66\stdvun{0.74} & \underline{45.15}\stdvun{0.88} & 86.40\stdvuru{1.88} & 67.38\stdvun{2.17} & 93.51\stdvun{0.12} & 21.93\stdvuru{1.89} \\

RLCF${\color{cyan}_{\text{ICLR '24}}}$ & 68.13\stdvun{0.21} & 45.15\stdvun{0.88} & 86.56\stdvuru{1.72} & 66.96\stdvun{1.75} & \underline{94.04}\stdvun{0.65} & 21.51\stdvuru{2.31} \\ 

\rowcolor[HTML]{E2EFDA} 
\(\text{TNT}*\) & 66.26\stdvuru{1.66} & 43.85\stdvuru{0.42} & 87.84\stdvuru{0.44} & 67.89\stdvun{2.68} & 93.31\stdvuru{0.08} & \textbf{24.54}\stdvun{0.72} \\

\rowcolor[HTML]{E2EFDA} 
\(\text{TNT}\) & 68.74\stdvun{0.82} & 41.71\stdvuru{2.56} & \textbf{89.59}\stdvun{1.31} & \textbf{68.81}\stdvun{3.60} & \textbf{94.66}\stdvun{1.27} & 20.65\stdvuru{3.17} \\ \hline

\end{tabular}
}

\vspace{0.01cm}

\centering
\resizebox{\textwidth}{!}{%
\begin{tabular}{l|cccc|c} \hline

\textbf{Method} $\downarrow$ & ~~~\textbf{EuroSAT}\cite{helber2019eurosat}~~~ & ~~~\textbf{StanfordCars}\cite{krause20133d}~~~ & ~~~\textbf{Food101}\cite{bossard2014food}~~~ & ~~~\textbf{SUN397}\cite{xiao2010sun}~~~ & ~~~~~\textbf{Average}~~~~~ \\ 
\cmidrule(r){1-1} \cmidrule(lr){2-2} \cmidrule(r){3-3} \cmidrule(r){4-4} \cmidrule(r){5-5} \cmidrule(l){6-6} 

CLIP-ViT-B/16 & 42.02\stdvuno{$bs.$} & 65.59\stdvuno{$bs.$} & 83.65\stdvuno{$bs.$} & 62.61\stdvuno{$bs.$} & 63.68\stdvuno{$bs.$} \\

CoOp & 41.78\stdvuru{0.24} & 64.74\stdvuru{0.85} & 83.96\stdvun{0.31} & 64.67\stdvun{2.06} & 63.20\stdvuru{0.48} \\

TPT${\color{cyan}_{\text{NIPS '22}}}$ & 42.81\stdvun{0.79} & 66.57\stdvun{0.98} & \underline{84.47}\stdvun{0.82} & \underline{64.82}\stdvun{2.21} & 64.18\stdvun{0.50} \\

CALIP${\color{cyan}_{\text{AAAI '23}}}$ & 42.27\stdvun{0.25} & 66.50\stdvun{0.91} & \textbf{86.93}\stdvun{3.28} & 63.48\stdvun{0.87} & 64.05\stdvun{0.37} \\

C-TPT${\color{cyan}_{\text{ICLR '24}}}$ & 36.33\stdvuru{5.69} & 64.77\stdvuru{0.82} & 81.67\stdvuru{2.98} & 62.56\stdvuru{0.05} & 62.86\stdvuru{0.82} \\

SaLS${\color{cyan}_{\text{ECCV '24}}}$ & 42.53\stdvun{0.51} & 66.73\stdvun{1.14} & 84.47\stdvun{0.82} & 64.81\stdvun{2.20} & 64.16\stdvun{0.48} \\

RLCF${\color{cyan}_{\text{ICLR '24}}}$ & \textbf{45.28}\stdvun{3.26} & 65.85\stdvun{0.26} & 84.06\stdvun{0.41} & 64.59\stdvun{1.98} & \underline{64.21}\stdvun{0.53} \\

\rowcolor[HTML]{E2EFDA} 
\(\text{TNT}*\) & 41.99\stdvuru{0.03} & \underline{66.93}\stdvun{1.34} & 84.10\stdvun{0.45} & 64.03\stdvun{1.42} & 64.07\stdvun{0.39} \\

\rowcolor[HTML]{E2EFDA} 
\(\text{TNT}\) & \underline{44.31}\stdvun{2.29} & \textbf{67.38}\stdvun{1.79} & 83.94\stdvun{0.29} & \textbf{65.00}\stdvun{2.39} & \textbf{64.48}\stdvun{0.80} \\ \hline

\end{tabular}
}
\label{tab:cross_data}
\vspace{-0.3cm}
\end{table*}
}

\section{Analysis and Ablations}

We conduct extensive analysis and ablations to assess how design choices impact performance, using the ImageNet-A benchmark with the ViT-B/16 backbone for consistency, as it represents a basic domain generalization variant.

\begin{figure}[t]
    \includegraphics[width=0.48\textwidth]{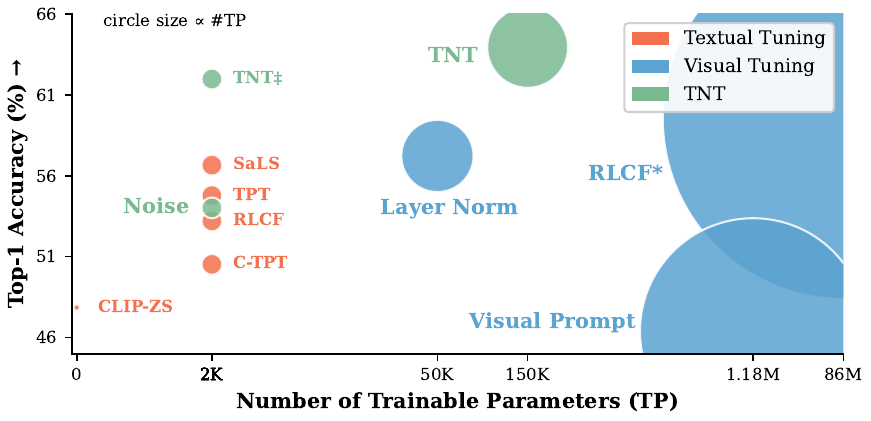}
     \vspace{-0.6cm}
     \caption{\textbf{Analysis of Trainable Parameters (TP)} for TNT, textual tuning, and encoder tuning. Circle size indicates the \#TP. {\color{customrocket}{Textual Tuning methods}} use the same TP count of 2K to optimize prompts. {\color{customblue}RLCF\textsuperscript{*}} refers to RLCF with all visual encoder parameters trainable, {\color{customblue}{Layer Norm}} limits trainable parameters of visual encoder to only Layer Norms, and {\color{customblue}{Visual Prompt}} applies learnable prompts to the visual encoder across 12 layers of the ViT encoder. {\color{customgreen}{TNT$\ddag$}} indicates TNT with only \(224\times9\) trainable noise parameters, compared to standard {\color{customgreen}{TNT}} with \(224\times224\times3\) TP. {\color{customgreen}{Noise}} denotes optimization with \(224\times9\) TP in noise and with only \(\mathcal{L}_{\text{entropy}}\) loss.  
     }
    \label{fig:params}
    \vspace{-2em}
\end{figure}


\subsection{Computational Analysis}
\label{comp_analysis}
\noindent\textbf{Trainable Parameters (TP) vs. Accuracy:}
\label{sec:train_param}
Test-time tuning methods like TPT, C-TPT, SaLS, and RLCF (prompt-tuning variant) adapt textual prompts with only 2048 TP (4 tokens of d=512) but show limited generalization, with Top-1 ImageNet-A accuracy around 50\% to 55\%. In contrast, visual adaptation based methods \textit{e.g.}  encoder tuning, visual prompting, and layer norm optimization—require more TP.
As depicted in Figure \ref{fig:params}, these Visual Tuning approaches offer moderate to lower generalization with increased TP, and that too while accessing the encoder itself.
In contrast, TNT effectively balances this trade-off with 150k (\(224\times224\times3\)) TP, achieving the highest generalization at 63.93, \textit{while preserving the black-box assumption}. Remarkably in our ablation, when TNT is initialized with 2016 (\(224\times9\)) TP (similar to the parameter count of textual tuning methods), it still demonstrates stronger generalization than baselines, highlighting its adaptability in handling distribution shifts through noise tuning. Furthermore, when TNT’s noise component (with 2016 TP) is tuned using only entropy (without \(\mathcal{L}_{\text{vc}}\) or temperature-scaled inference), it achieves performance on par with that of textual tuning approaches.

\vspace{0.2cm}
\noindent\textbf{Trade-off between Accuracy, Time, and Memory:} Table \ref{tab:computation} shows that TNT achieves the highest performance while maintaining optimal inference time and memory efficiency if not the best. 
Although TNT demands a moderate memory increase (8.57 GB), it maintains practical inference speed and accuracy compared to alternatives like C-TPT, which incurs higher memory (up to 33.44 GB for ImageNet-V) and slower inference times. However, our approach demonstrates the best trade-off by achieving top accuracy with efficient memory use, especially in cases where feature complexity necessitates higher resource allocation for robust predictions. 

{\renewcommand{\arraystretch}{1.3}
\begin{table}[t]
\setlength{\tabcolsep}{4pt}
\centering
\caption{\textbf{Average GPU inference time per sample and memory usage} across different optimization steps. Top-1 Accuracy (Acc), inference time in seconds (Time), and memory usage in GB (Mem) are shown for each method. \textbf{Bold} indicates best performance.}
\label{tab:computation}
\fontsize{12}{12}\selectfont
\resizebox{0.475\textwidth}{!}{%
\begin{tabular}{l|c|ccc|ccc}
\hline

\multirow{2}{*}{\textbf{Method} $\downarrow$} & \multirow{2}{*}{\textbf{Steps}} & \multicolumn{3}{c|}{\textbf{ImageNet-A}} & \multicolumn{3}{c}{\textbf{ImageNet-V}} \\
\cmidrule(lr){3-5} \cmidrule(l){6-8}
 &  & \multicolumn{1}{c}{\textbf{Acc}} & \multicolumn{1}{c}{\textbf{Time}} & \textbf{Mem} & \multicolumn{1}{c}{\textbf{Acc}} & \multicolumn{1}{c}{\textbf{Time}} & \textbf{Mem} \\
\cmidrule(r){1-1} \cmidrule(lr){2-2} \cmidrule(lr){3-3} \cmidrule(lr){4-4} \cmidrule(lr){5-5} \cmidrule(lr){6-6} \cmidrule(lr){7-7} \cmidrule(l){8-8}
 
CLIP-ViT-B/16 & 0 & \multicolumn{1}{c}{47.85} & \multicolumn{1}{c}{0.05} & 2.06 & \multicolumn{1}{c}{60.89} & \multicolumn{1}{c}{0.20} & 3.73 \\  
\cmidrule(r){1-1} \cmidrule(lr){2-2} \cmidrule(lr){3-3} \cmidrule(lr){4-4} \cmidrule(lr){5-5} \cmidrule(lr){6-6} \cmidrule(lr){7-7} \cmidrule(l){8-8}

\multirow{2}{*}{TPT${\color{cyan}_{\text{NIPS '22}}}$} & 1 & \multicolumn{1}{c}{54.31} & \multicolumn{1}{c}{\textbf{0.21}} & \textbf{5.05} & \multicolumn{1}{c}{63.19} & \multicolumn{1}{c}{\textbf{0.71}} & \textbf{18.35} \\ 
 & 3 & \multicolumn{1}{c}{57.91} & \multicolumn{1}{c}{\textbf{0.50}} & \textbf{5.05} & \multicolumn{1}{c}{65.02} & \multicolumn{1}{c}{\textbf{1.76}} & \textbf{18.35} \\ 
 \cmidrule(r){1-1} \cmidrule(lr){2-2} \cmidrule(lr){3-3} \cmidrule(lr){4-4} \cmidrule(lr){5-5} \cmidrule(lr){6-6} \cmidrule(lr){7-7} \cmidrule(l){8-8}
 
\multirow{2}{*}{C-TPT${\color{cyan}_{\text{ICLR '24}}}$} & 1 & \multicolumn{1}{c}{50.52} & \multicolumn{1}{c}{0.36} & 8.19 & \multicolumn{1}{c}{61.56} & \multicolumn{1}{c}{1.23} & 33.44 \\
 & 3 & \multicolumn{1}{c}{54.42} & \multicolumn{1}{c}{0.96} & 8.19 & \multicolumn{1}{c}{64.72} & \multicolumn{1}{c}{3.32} & 33.44 \\ 
 \cmidrule(r){1-1} \cmidrule(lr){2-2} \cmidrule(lr){3-3} \cmidrule(lr){4-4} \cmidrule(lr){5-5} \cmidrule(lr){6-6} \cmidrule(lr){7-7} \cmidrule(l){8-8}
 
\multirow{2}{*}{SaLS${\color{cyan}_{\text{ECCV '24}}}$} & 1 & \multicolumn{1}{c}{56.65} & \multicolumn{1}{c}{0.21} & 5.05 & \multicolumn{1}{c}{63.22} & \multicolumn{1}{c}{0.71} & 18.35 \\
 & 3 & \multicolumn{1}{c}{58.16} & \multicolumn{1}{c}{0.51} & 5.05 & \multicolumn{1}{c}{64.57} & \multicolumn{1}{c}{1.76} & 18.35 \\
 \cmidrule(r){1-1} \cmidrule(lr){2-2} \cmidrule(lr){3-3} \cmidrule(lr){4-4} \cmidrule(lr){5-5} \cmidrule(lr){6-6} \cmidrule(lr){7-7} \cmidrule(l){8-8}
 
\multirow{2}{*}{RLCF${\color{cyan}_{\text{ICLR '24}}}$} & 1 & \multicolumn{1}{c}{54.77} & \multicolumn{1}{c}{0.22} & 5.05 & \multicolumn{1}{c}{62.71} & \multicolumn{1}{c}{0.73} & 18.35 \\
 & 3 & \multicolumn{1}{c}{57.27} & \multicolumn{1}{c}{0.52} & 5.05 & \multicolumn{1}{c}{64.02} & \multicolumn{1}{c}{1.78} & 18.35 \\
 \cmidrule(r){1-1} \cmidrule(lr){2-2} \cmidrule(lr){3-3} \cmidrule(lr){4-4} \cmidrule(lr){5-5} \cmidrule(lr){6-6} \cmidrule(lr){7-7} \cmidrule(l){8-8}


\rowcolor[HTML]{E2EFDA} 
\cellcolor[HTML]{E2EFDA} & 1 & \multicolumn{1}{c}{\cellcolor[HTML]{E2EFDA}\textbf{63.93}} & \multicolumn{1}{c}{\cellcolor[HTML]{E2EFDA}0.33} & 8.57 & \multicolumn{1}{c}{\cellcolor[HTML]{E2EFDA}\textbf{66.64}} & \multicolumn{1}{c}{\cellcolor[HTML]{E2EFDA}0.87} & 21.89 \\
\rowcolor[HTML]{E2EFDA} 
\multirow{-2}{*}{\cellcolor[HTML]{E2EFDA}\textbf{TNT}} & 3 & \multicolumn{1}{c}{\cellcolor[HTML]{E2EFDA}\textbf{65.17}} & \multicolumn{1}{c}{\cellcolor[HTML]{E2EFDA}0.79} & 8.57 & \multicolumn{1}{c}{\cellcolor[HTML]{E2EFDA}\textbf{67.10}} & \multicolumn{1}{c}{\cellcolor[HTML]{E2EFDA}2.09} & 21.89 \\ \hline

\end{tabular}}
\vspace{-0.4cm}
\end{table}}

\subsection{Ablations}\label{ablations}
\begin{figure}[t]
    \centering
    \begin{subfigure}[b]{0.45\textwidth}
        \includegraphics[width=\textwidth]{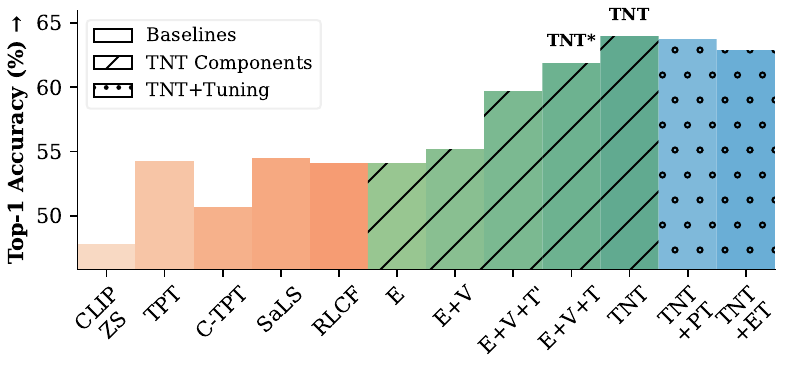}
        \vspace{-0.6cm}
        \caption{Top-1 Accuracy}
    \end{subfigure}
    \begin{subfigure}[b]{0.45\textwidth}
        \includegraphics[width=\textwidth]{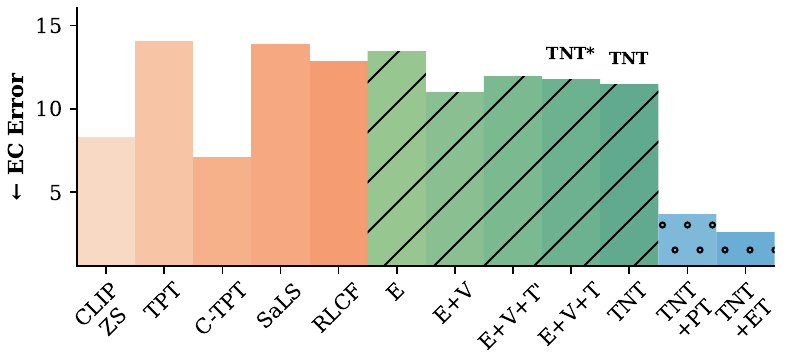}
        \vspace{-0.6cm}
        \caption{Expected Calibration Error \cite{yooncctpt}}
    \end{subfigure}
    \caption{
    \textbf{Effect of TNT Components on (a) Top-1 Accuracy (Higher$\uparrow$ is better) and (b) ECE (Lower$\downarrow$ is better)}.
    \textbf{E}: Noise optimization with Entropy minimization $\mathcal{L}_{\text{entropy}}$.
    \textbf{E+V}: Adds $\mathcal{L}_{\text{vc}}$ loss (Eq.~\ref{eq:Lvc_loss}) to $\mathcal{L}_{\text{entropy}}$.
    \textbf{E+V+T$^\prime$}: Adds temperature scaling during inference to E+V. 
    \textbf{E+V+T}: Makes use of top-$K$ views instead of one test image (Eq.~\ref{eq:inference}), i.e. TNT*. 
    \textbf{TNT}: TNT* with CoOp initialization, 
    \textbf{TNT+PT}(\textit{Prompt Tuning}): Optimizes textual prompts with TNT. 
    \textbf{TNT+ET}(\textit{Encoder Tuning}): Optimizes the visual encoder with TNT. Optimization Steps $t=1$ is used consistently. The same Legend is used for (a) and (b).
    }
    \label{fig:different_components}
    \vspace{-0.50cm}
\end{figure}

\noindent\textbf{Effect of Different Components:}
Although TNT's noise tuning shows enhanced results, it is intriguing to understand how each component contributes to improved generalization and calibration. As depicted in Figure \ref{fig:different_components}, initialized noise, when optimized with only entropy (\textbf{E}), shows comparable generalization and calibration to textual tuning baselines. Adding the inter-view consistency loss (\textbf{E+V}) improves the alignment of image embeddings, resulting in more consistent and confident predictions. This approach not only outperforms entropy minimization alone but also reduces the Expected Calibration (EC) error by 3\%. Intuitively, this process encourages the embeddings to be closer in feature space, leading to lower intra-class variance while preserving inter-class differences. 
Furthermore, during inference, applying temperature \(\tau\) to scale the output logits (Eq.~\ref{eq:inference}), as in \textbf{E+V+T$^\prime$}, increases accuracy by 3\%. Extending this approach to consider the top-\(K\) views, as in \textbf{E+V+T}, further improves performance by 2\%. This configuration corresponds to our TNT* variant. The top-\(K\) strategy focuses final predictions on the most confident views rather than relying solely on a single sample, resulting in notable accuracy gains.
When TNT* is initialized with CoOp, resulting in \textbf{TNT}, it achieves the highest generalization (63.93 Top-1 accuracy) and reliable calibration error \cite{yooncctpt, murugesan2024robust}. TNT achieves an ECE of 11.46, second only to C-TPT, while outperforming all other baselines in accuracy. This result indicates strong uncertainty estimation with effective noise tuning.

\vspace{0.2cm}
\noindent\textbf{Impact of Combinative Tuning:} Assumably, when combining the proposed TNT with prompt tuning (PT) or encoder tuning (ET), one might expect better generalization. Interestingly, we observe in Figure \ref{fig:different_components} that both \textbf{TNT+PT} and \textbf{TNT+ET} achieve comparable generalization to TNT while \textit{significantly improving calibration error}. Intuitively, this is because \textbf{TNT+PT} refines the alignment between textual and visual features, enhancing calibration. On the other hand, \textbf{TNT+ET} improves calibration by allowing the model to better align its learned visual representations with the specific distribution of the input data. This shows that calibration error can be further minimized by optimizing the interaction between the model's visual and textual components, on top of the TNT framework.

\begin{figure}[t]
    \centering
    \begin{subfigure}[b]{\linewidth}
        \includegraphics[width=0.49\textwidth]{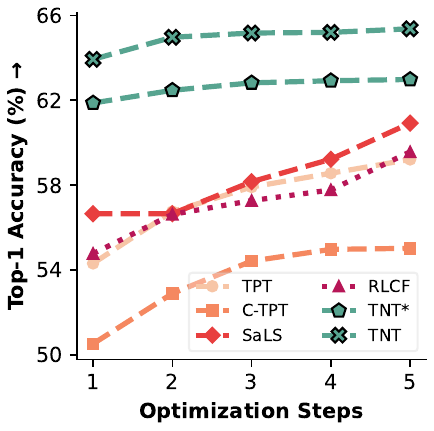}
        \includegraphics[width=0.49\textwidth]{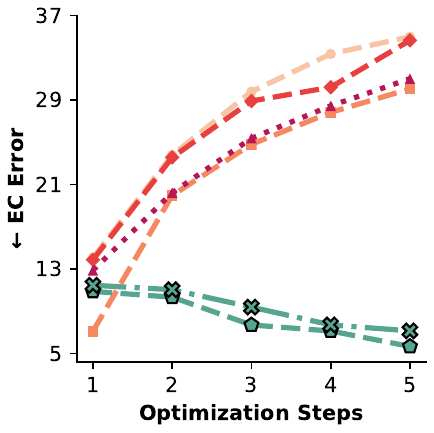}
        \caption{Effect of Number of Optimization Steps (with 64 augmentations)}
    \end{subfigure}
    \begin{subfigure}[b]{\linewidth}
            \includegraphics[width=0.49\textwidth]{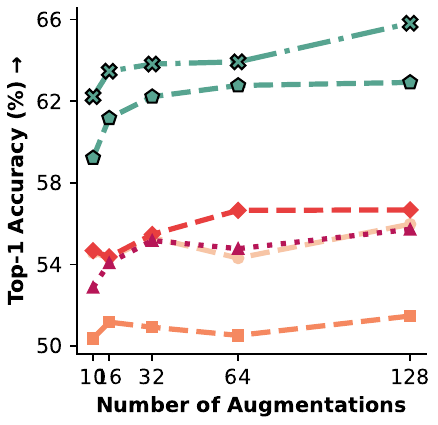}
        \includegraphics[width=0.49\textwidth]{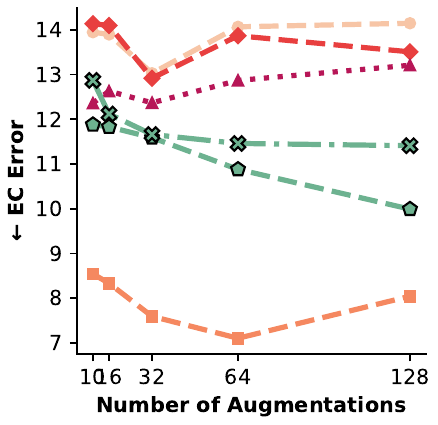}
        \caption{Effect of Number of Augmentations (with 1 optimization step)}
    \end{subfigure}
    \vspace{-0.6cm}
    \caption{Increasing the number of optimization steps and augmentations both result in higher \textbf{(a) Top-1 Accuracy}, and lower \textbf{(b) Expected Calibration (EC) Error}. TNT* and TNT denote hand-crafted and CoOp-based prompts, respectively. The legend is shared throughout.}
    \label{fig:ablations}
    \vspace{-0.55cm}
\end{figure}


\vspace{0.2cm}
\noindent\textbf{Effect of Optimization Steps:}
All reported results thus far have been obtained with a single training step; however, we observe that TNT’s efficacy is influenced by the number of optimization steps, impacting both accuracy and ECE \cite{yooncctpt}. In Figure \ref{fig:ablations}(a)(left), we observe that increasing the number of steps leads to a 2\% performance gain for both TNT* and TNT configurations from step $t=1$ to $t=5$. While other baselines follow a similar trend across optimization steps, they exhibit lower accuracy compared to TNT, indicating TNT’s adaptability for more efficient applications. Interestingly, in Figure \ref{fig:ablations}(a)(right), we find that TNT variants achieve notably lower calibration error with more optimization steps, as TNT* reaches an ECE of 5.67 at step $t=5$, outperforming zero-shot CLIP (ECE 8.32) and other baselines by a considerable margin. The additional optimization steps in TNT encourage the noise to increasingly adapt to more representative features, as shown in Figure \ref{fig:concept}.

\vspace{0.2cm}
\noindent\textbf{Effect of Number of Augmentations:}
Similarly, accuracy improves with an increasing number of augmentations across TNT configurations and baselines, reaching a plateau at \(N=64\) for most baselines as shown in Figure \ref{fig:ablations}(b)(left). Notably, TNT achieves a Top-1 Accuracy of 65.82 with 128 augmentations, though with higher memory requirements. In contrast, TNT* sees a modest 0.20\% increase from \(N=64\) to \(N=128\) augmentations. Regarding ECE, TNT also reduces ECE with an increasing number of augmentation steps, where TNT* achieves the second-best ECE at 9.99, following C-TPT at 8.04, while TPT reaches 14.15 as shown in Figure \ref{fig:ablations}(b)(right). Varied augmentations provide more \textit{diverse} and class-specific crops, leading to better overall generalization and lower EC error. Given the linear increase in memory usage associated with additional augmentations, we adopt single-step optimization as the default setting for both TNT* and TNT, as they already improve the average OOD accuracy over zero-shot CLIP by 5.42\% and 7.38\% respectively. 



\subsection{Qualitative Analysis}
\noindent\textbf{Impact of Noise Optimization on Attention Maps:}
TNT optimizes noise in a way that it \textit{implicitly} influences the input’s attention maps, guiding the model to adaptively focus on the most relevant features within the test input. This process enables TNT to dynamically emphasize important regions, ensuring the model attends to critical information for each sample. As illustrated in Figure \ref{fig:concept}, there is a noticeable refinement in the attention between the original and adapted samples; with each optimization step, the adapted noise becomes increasingly \textit{aligned} with the salient features, effectively suppressing irrelevant details and amplifying relevant cues. This progressive adaptation results in more confident and accurate predictions, as the model’s focus narrows on features that are contextually meaningful and better aligned with the target task.

\begin{figure}[t]
    \centering    
    \begin{subfigure}[b]{0.155\textwidth}
        \includegraphics[width=\textwidth]{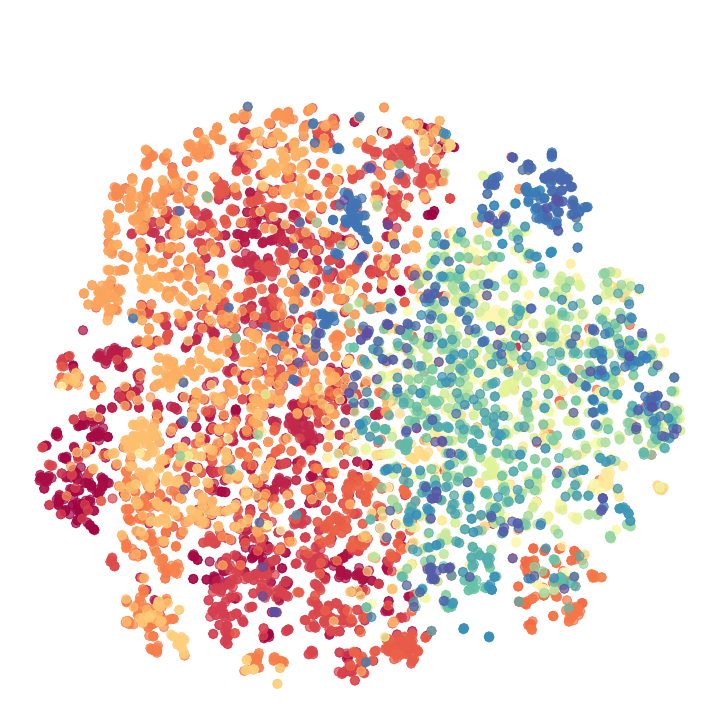}
        \caption{CLIP-ZS \cite{radford2021learning}}
    \end{subfigure}
    \begin{subfigure}[b]{0.155\textwidth}
        \includegraphics[width=\textwidth]{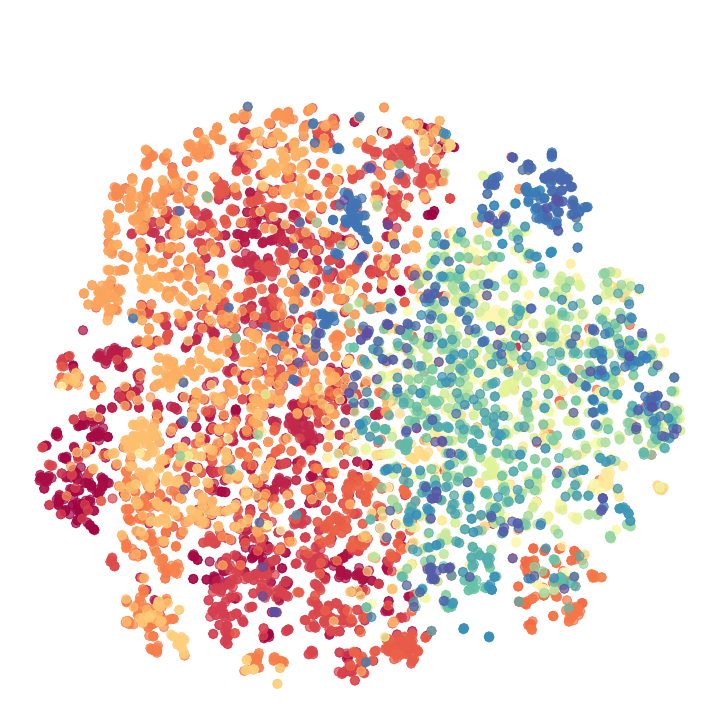}
        \caption{TPT \cite{shu2022test}}
    \end{subfigure}
    \begin{subfigure}[b]{0.155\textwidth}
        \includegraphics[width=\textwidth]{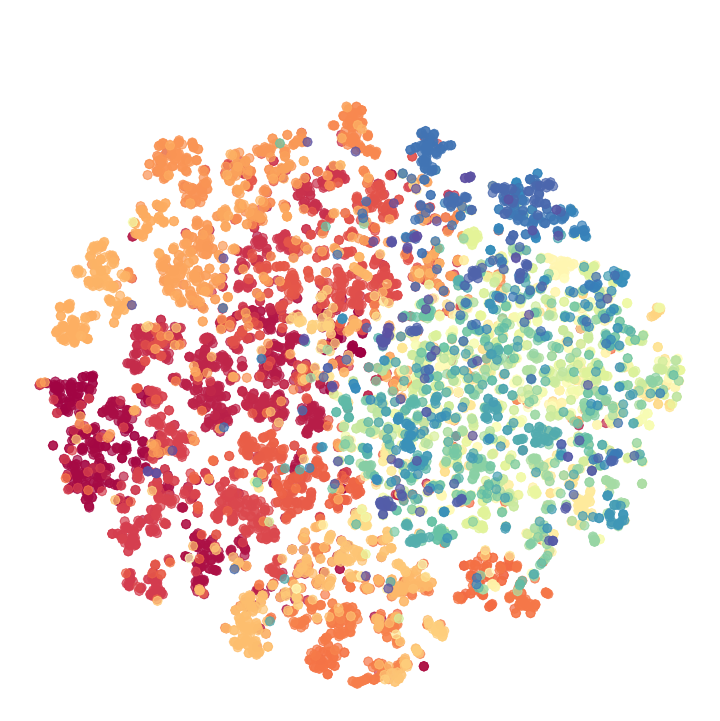}
        \caption{\textbf{TNT} (Ours)}
    \end{subfigure}
    \caption{\textbf{t-SNE visualizations} of the final class embedding from the test sets of ImageNet-A dataset, following Table \ref{tab:natural_shift}. TNT could produce \textit{more clustered and separable features} than other zero-shot generalization baselines.}
    \label{fig:tsne}
    \vspace{-0.5cm}
\end{figure}

\vspace{0.2cm}
\noindent\textbf{Feature Shifts via Adaptation:}
Figure \ref{fig:tsne} examines the shift in the distribution of visual features after optimization, comparing the results from baselines and TNT.
As depicted, CLIP-ZS and TPT show scattered feature distributions, reflecting their struggle to distinguish class boundaries effectively in naturally shifted data. In contrast, TNT demonstrates tightly \textit{clustered} and \textit{well-separated} features, which suggests that TNT’s noise adaptation mechanism and consistency losses promote feature alignment and separation more effectively than entropy-based or prompt-tuning methods alone. 
This focused feature adaptation helps establish more \textit{defined decision boundaries} and mitigates the impact of irrelevant features, thus leading to better overall performance on out-of-distribution samples.


\section{Conclusion}
We introduce Test-Time Noise Tuning (TNT), a novel noise adaptation strategy for zero-shot settings that enhances out-of-distribution generalization by tuning a learnable noise for visual input of a VLM, improving robustness and calibration. TNT demonstrates the potential of noise tuning in challenging VLM benchmarks, setting a foundation for adaptive OOD handling. Our noise tuning demonstrates, for the first time, a positive impact on representation learning in VLMs, paving the way for further exploration across other modalities. Extending TNT’s noise tuning and inter-view consistency loss to other vision-language tasks, such as retrieval, as well as applications like medical imaging, would be valuable.
A promising direction for future research is to explore strategies for reducing the memory requirements of TNT, enhancing its scalability and applicability in resource-constrained environments.

{
\small
\bibliographystyle{ieeenat_fullname}
\bibliography{main}
}


\end{document}